\pdfoutput=1

\documentclass[11pt]{article}

\usepackage[preprint]{coling}

\usepackage{times}
\usepackage{latexsym}

\usepackage[T1]{fontenc}

\usepackage[utf8]{inputenc}

\usepackage{microtype}

\usepackage{inconsolata}

\usepackage{graphicx}
\usepackage{booktabs}

\usepackage{enumitem}

\title{Automated Molecular Concept Generation and Labeling with Large Language Models}

\author{
 \textbf{Zimin Zhang\textsuperscript{1}}\footnotemark[1]\footnotemark[2],
 \textbf{Qianli Wu\textsuperscript{2}\footnotemark[1]\footnotemark[2]},
 \textbf{Botao Xia\textsuperscript{2}\footnotemark[1]\footnotemark[2]},
\\
 \textbf{Fang Sun\textsuperscript{2}},
 \textbf{Ziniu Hu\textsuperscript{3}},
 \textbf{Yizhou Sun\textsuperscript{2}},
 \textbf{Shichang Zhang\textsuperscript{4}}\footnotemark[2],
\\
 \textsuperscript{1}University of Illinois Urbana-Champaign,
 \textsuperscript{2}University of California Los Angeles \\
 \textsuperscript{3}California Institute of Technology,
 \textsuperscript{4}Harvard University
\\
 \textsuperscript{1}\texttt{ziminz19@illinois.edu}\\
 \textsuperscript{2}\texttt{\{qianliwu, xiabotao\}@g.ucla.edu, \{fts, yzsun\}@cs.ucla.edu}\\
 \textsuperscript{3}\texttt{acgbull@gmail.com}, \textsuperscript{4}\texttt{shzhang@hbs.edu}
}

\newcommand{\method}{\textmd{AutoMolCo}}

\begin{document}
\renewcommand{\thefootnote}{\fnsymbol{footnote}}
\maketitle
\footnotetext[1]{Equal Contribution.}
\footnotetext[2]{Work done when authors were at University of California Los Angeles.}

\begin{abstract}
Artificial intelligence (AI) is transforming scientific research, with explainable AI methods like concept-based models (CMs) showing promise for new discoveries. However, in molecular science, CMs are less common than black-box models like Graph Neural Networks (GNNs), due to their need for predefined concepts and manual labeling. This paper introduces the \textbf{Automated Molecular Concept (AutoMolCo)} framework, which leverages Large Language Models (LLMs) to automatically generate and label predictive molecular concepts. Through iterative concept refinement, AutoMolCo enables simple linear models to outperform GNNs and LLM in-context learning on several benchmarks. The framework operates without human knowledge input, overcoming limitations of existing CMs while maintaining explainability and allowing easy intervention. Experiments on MoleculeNet and High-Throughput Experimentation (HTE) datasets demonstrate that AutoMolCo-induced explainable CMs are beneficial for molecular science research. The source code is available at \url{https://github.com/ziminz19/AutoMolCo}.
\end{abstract}

\section{Introduction}
\label{sec:introduction}
Artificial intelligence (AI) has significantly advanced molecular science. A prime example is MIT Jameel Clinic's use of deep learning to identify halicin – the first antibiotic discovered in three decades that is effective against a broad spectrum of 35 bacteria~\cite{stokes2020deep}. Deep learning models, such as Graph Neural Networks (GNNs), excel at learning complex atomic structures and predicting molecular properties~\cite{wu2018moleculenet}.  However, a major challenge with such deep-learning-based models like GNNs is their ``black boxes'' nature and lack of explainability~\cite{yuan2022explainability}. Despite their high predictive performance, black-box models fail to provide insights into the underlying reasoning behind their predictions, making it difficult for scientists to interpret and intervene in the model's decision-making process, which hinders scientific understanding and limits the potential for knowledge discovery.

\begin{figure*}[h]
\centering
\includegraphics[width=\textwidth]{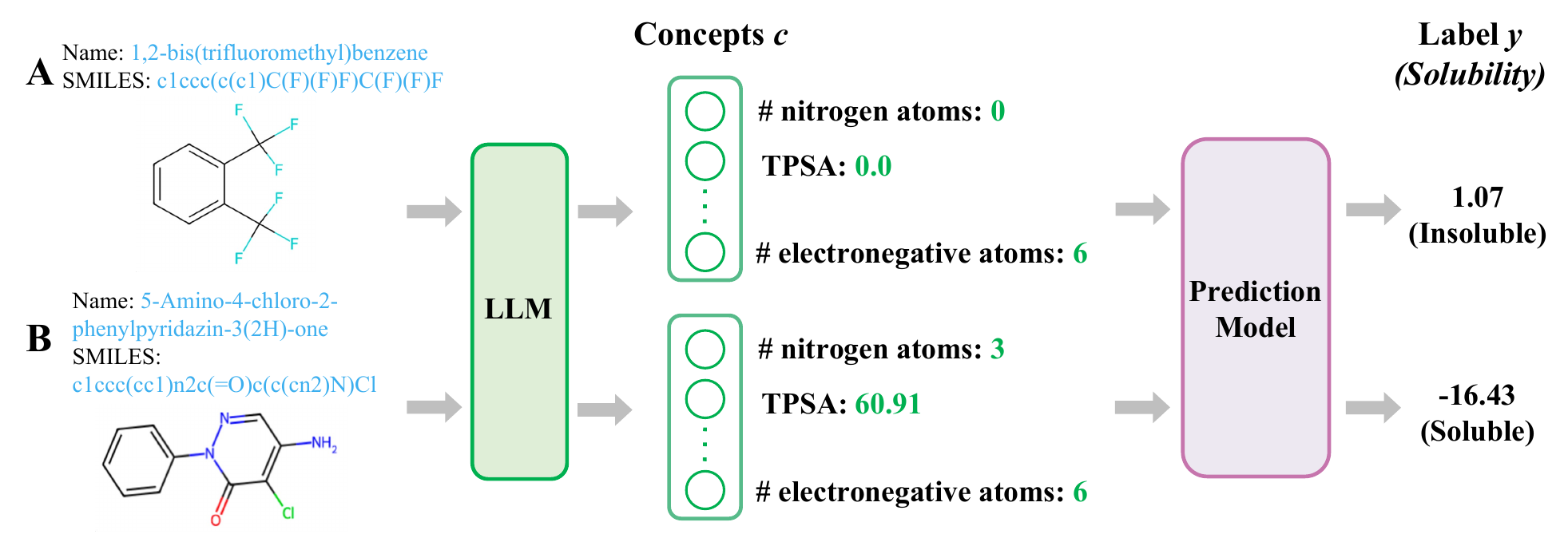} 
\caption{The prediction process of molecule properties is greatly illuminated with \method. First, concepts are generated and labeled with an LLM. Then, a simple prediction model (e.g., linear regression) is fitted to achieve explainable predictions. LLM-relevant pieces are highlighted in green.}
\vskip -0.2in
\label{figure:hook}
\end{figure*}

In contrast, concept-based models (CMs)~\citep{lampert2009learning, koh2020concept, yeh2020completeness, wu2023causal} offer a promising explainable AI (XAI) approach by providing insights that can drive scientific discoveries. Unlike black-box models, CMs first predict human-interpretable concepts and then use them to predict task labels, providing both predictions and rationales. For example, in computer vision, CMs predict bird species by identifying concepts like "wing color"~\cite{koh2020concept}. In molecular science, CMs interpret predictions through concepts like functional groups and molecular descriptors. As shown in Figure~\ref{figure:hook}, a CM predicts molecular solubility using descriptors like \textit{\# of nitrogen atoms} and \textit{TPSA}, allowing researchers to refine molecules based on these key features.

Despite their promise, CMs have seen limited application in molecular science due to challenges in concept generation and labeling. Existing CM methods either rely on predefined concepts and manual labeling by experts~\cite{koh2020concept} or are limited to simple, qualitative concepts inadequate for molecular problems~\cite{oikarinen2023label}. While feasible in computer vision~\cite{koh2020concept, oikarinen2023label}, molecular concepts are more complex and require precise quantitative labels, like TPSA in Figure~\ref{figure:hook}, which reflects absorption and permeability relevant to solubility prediction. Identifying such concepts demands domain expertise and computational methods beyond current CM capabilities, posing a significant challenge for their effective use in molecular science.

In response to the effectiveness of CMs and the challenges of applying them to molecular science, we propose \textbf{Automated Molecular Concept (AutoMolCo)} generation and labeling. AutoMolCo leverages Large Language Models (LLMs) to generate molecular concepts that are predictive for the task and label these concepts for each molecule instance. AutoMolCo also repeats these procedures through iterative interactions with LLMs to refine concepts, enabling simple linear models on the refined concepts to outperform GNNs and LLM in-context learning (ICL) on several molecular benchmarks. The whole framework is automated and does not require human knowledge inputs in either concept generation, labeling, or refinement, thus surpassing the limitations of extant CMs.

The motivation behind AutoMolCo is the idea that LLMs can serve as extensive knowledge bases~\cite{petroni2019language, alkhamissi2022review}, with their effectiveness for solving molecular science problems demonstrated through ICL~\cite{guo2023can}. We leverage LLMs for XAI by integrating them into CMs. For concept generation, we prompt LLMs with the task description to suggest relevant concepts. For concept labeling, we explore three methods: direct LLM prompting, function code generation, and external tool calling. We then build simple prediction models on these concepts. Additionally, we iteratively refine concepts by running feature selection and prompting LLMs to generate improved concepts, ensuring the CM remains up-to-date with the most relevant concepts and enhances performance.

In this work, we first show that \method\ can produce meaningful concepts and accurate labels, which lead to CMs with simple prediction models to achieve surprisingly good performance for molecular science problems. Then we perform a systematic study of \method\ on MoleculeNet~\cite{wu2018moleculenet} and High-Throughput Experimentation (HTE)~\cite{bhpaper, suzuki} datasets to answer five research questions.  
In summary, our contribution includes:
\begin{enumerate}[nosep, wide=0pt, leftmargin=*, after=\strut, label=\textbullet]
    \item \textbf{Automated framework:} We propose \method, which leverages LLMs for automated concept generation and labeling, eliminating the need for human domain knowledge and labor-intensive data collection, thereby streamlining the development of CMs. 

    \item \textbf{Accuracy and explainability:} \method\ produces meaningful molecular concepts that, when combined with simple prediction models for CMs, can achieve superior or comparable accuracy to powerful black-box models while providing greater explainability.

    \item \textbf{LLM-driven XAI for science:} Our work highlights the potential of LLMs in addressing complex molecular science problems,
    introduces a novel perspective on CMs with LLMs, and paves the way for future research to exploit the LLMs' capabilities in molecular science and beyond.    
\end{enumerate}


\section{Related work}
\label{sec:related}
\textbf{Concept-based Models.} A well-known example of CMs is the Concept Bottleneck Model (CBM)~\cite{koh2020concept}, which predicts through an intermediate layer of human-specified concepts, like "wing color" in bird classification. While transparent, CBMs are limited by predefined concepts and label requirements. Several variations of CBMs target specific tasks~\cite{de2018clinically, yi2018neural, bucher2019semantic, losch2019interpretability, chen2020concept}, with a notable one, label-free CBM~\cite{oikarinen2023label}, bypasses predefined concepts by using GPT-3 for concept generation and CLIP-Dissect for matching concepts with images. However, it focuses on vision tasks and generates simple, qualitative concepts (e.g.,``yellow'') that are insufficient for molecules, which demand deeper chemical knowledge and precise quantitative labels.

\textbf{Explainable Learning on Scientific Graphs Data} Explainable learning on graph data is getting popular, especially for scientific problems like particle identifying~\cite{mokhtar2022graph}, whether prediction~\cite{jeon2024cloudnine}, material design~\cite{wang2020predicting,li2024predicting}, and in particular, molecular science~\cite{yuan2021explainability, zhang2022gstarx, wu2023chemistry}. One line identifies graph motifs as concepts through counting or sampling~\cite{milo2002network,wernicke2006efficient} and builds GNNs on top of them~\cite{zhang2020motif, yu2022molecular}. However, motif identification cannot be comprehensive as it is NP-complete. Another line tries to use concept-based explanations for GNNs with human-in-the-loop~\cite{magister2021gcexplainer}. Subsequent works have refined this idea with k-means clustering and similarity scoring algorithms to neuron-level grouping within activation layers~\cite{magister2022encoding, xuanyuan2023global}. These methods exemplify the attempt to extract and interpret salient features in graph data, yet they often face challenges in fully capturing the nuanced complexity of molecular structures.

\textbf{LLMs for Molecular Science.}
Recently, there are some benchmarking papers on LLMs for molecular science. 
GPT4Graph~\cite{guo2023gpt4graph} prompts LLMs to explain the format or to summarize a raw molecule graph input, where the graph is represented by the Graph Modelling Language (GML)~\cite{himsolt1997gml} or Graph Markup Language (GraphML)~\cite{brandes2013graph}. Graph-ToolFormer~\cite{zhang2023graph} lets LLMs generate API calls to use external graph reasoning tools, which can be applied to molecule function reasoning problems. 
\cite{guo2023can} studies solving molecular problems with LLMs ICL. We show \method\ outperforms ICL and enjoys better explainability. Some survey papers discussing LLMs' potential for molecular science include: \cite{zhang2023artificial} from a scientific research perspective and~\cite{jin2023large} from an LLM for graph perspective, and \cite{yu2024llasmol} for fine-tuning LLMs.

\section{AutoMolCo: automated molecular concept generation and labeling} \label{sec:methodology}
\begin{figure*}
\centering
\includegraphics[width=\textwidth]{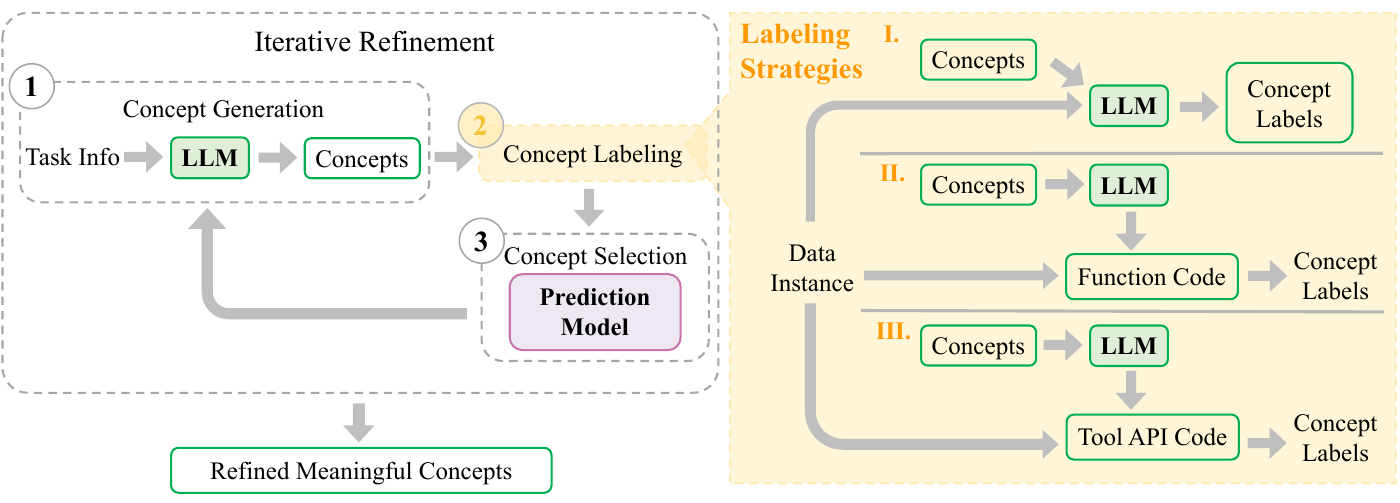} 
\caption[The \method\ framework.]{\textbf{The \method\ framework.} Step 1: concept generation. Step 2: concept labeling with three different strategies. Step 3: fitting a prediction model and perform concept selection. These three steps are repeated for multiple iterations to achieve a refined list of meaningful concepts, where the selected concepts in each iteration are feedback to the LLM through prompting. LLM outputs are highlighted as green boxes.}
\vskip -0.2in
\label{figure:framework}
\end{figure*}

In this section, we describe \method\ for concept generation, labeling, and refinement, where the concepts are used to build an explainable CM. Figure~\ref{figure:framework} depicts the three major steps of \method: 1) concept generation, 2) concept labeling, and 3) CM fitting and concept selection. 

\textbf{Step 1: Concept Generation}
Given a particular task on molecules, e.g., predicting the hydration-free energy of small molecules in water, the first step is to prompt LLMs to propose a diverse list of concepts that are potentially relevant to the task. This step is analogous to a brainstorming process. Concepts range from counting-based ones, like \textit{\# nitrogen atoms}, to more complicated ones that require precise calculation, like TPSA. Without LLMs, coming up with meaningful concepts requires domain experts. The underlying intuition for concept generation is founded on the idea that LLMs can be treated as extensive and integrated knowledge bases. Their capacity to comprehend and output meaningful concepts is pivotal in this phase, yielding a wide spectrum of potentially relevant concepts for our analysis. The prompt for this step is shown in Figure~\ref{figure:llm_prompt} Step 1. The LLM-suggested concepts might be less relevant initially, but they will be refined later.

\textbf{Step 2: Concept Labeling}
Following the concept generation step, we then label the generated concepts for each data instance. Compared to human labeling, which requires domain knowledge and can be labor-intensive. Labeling with LLMs is streamlined to a process of interaction with a single LLM interface, which can be easily scaled and minimizes human error. This automation with LLMs is crucial for efficiently processing large volumes of data encountered in molecular studies. In this step, we consider three different labeling strategies to enhance labeling quality. 

\textit{Labeling Strategy 1: Direct LLM prompting.}
    We prompt LLMs directly to assign each data instance numerical or categorical labels for the generated concepts from Step 1. Similar to concept generation, this strategy relies on that LLMs can be treated as integrated knowledge bases for retrieving useful information. For each data instance, we provide LLMs with the molecule names or SMILES strings. The prompt is shown in Figure~\ref{figure:llm_prompt} Step 2.

\textit{Labeling Strategy 2: Function code generation with LLMs.}
    Since LLMs are particularly skilled in code generation, we explore a second approach for concept labeling: generate functions in Python code for computing the concept labels. The function generation approach has two advantages. Firstly, it greatly reduces the need for repeated LLM API calls. Only a single API call is required for each concept to obtain the function code, as opposed to making a separate call for each data instance in the direct label prompting case. Secondly, the generated functions can utilize preprocessed dataset features as function arguments, such as atom types in terms of node features and molecule structures in terms of adjacency matrix. These features provide more direct information beyond molecule names or SMILES strings. Leveraging these features, the LLM-generated functions can offer more nuanced and accurate concept labels, enhancing the effectiveness of \method. The prompt is shown in Figure~\ref{figure:function_generation} in Appendix~\ref{app:prompts}. 

\textit{Labeling Strategy 3: External tool calling with LLMs.}
    We also utilize LLMs to call external tools like RDKit~\cite{rdkit} for labeling, which combines the LLM generation with the specialized tool reliability. This strategy enjoys the same efficiency advantage as the function generation approach, meaning it requires only a single API call of the LLM per concept to get the API code for calling the tool. 
    Moreover, the use of labeling tools ensures that labels for all the tool-calculable concepts are accurate and reliable. One disadvantage of this strategy is that not all generated concepts are calculable by the external tool, in which case we can only turn to the first two strategies. The prompt is shown in Figure~\ref{figure:external_tool} in Appendix~\ref{app:prompts}. 

\textbf{Step 3: CM Fitting and Concept Selection}
After getting the generated concepts and their labels, we utilize them to fit prediction models for the molecular task. Since the concept labels can be treated as tabular data, any model from the off-the-shelf ones in Scikit-learn~\cite{scikit-learn} to sophisticated deep learning models can be applied. However, we found that explainable models like linear models and decision trees or simple two-layer multi-layer perceptions (MLPs) are often sufficient for achieving competitive performance. We attribute the credit to the high-quality concepts and their labels. We will discuss our model choice and perform a systematic study of different prediction models in Section~\ref{sec:experiment}. While fitting the model, we also run feature selection methods like Akaike Information Criterion (AIC)~\cite{akaike1973information, akaike1974new} and Recursive Feature Elimination (RFE)~\cite{rfe} to determine the useful concepts. Feature selection not only boosts the model performance but also leads to automated iterative refinement for identifying the most useful concepts.

\textbf{Iterative Concepts Refinement}
After all three steps. We do an iterative refinement of the generated concepts by prompting LLMs again with the empirical performance of our prediction model and the concept selection results from Step 3. We include such information in an updated prompt to make LLMs generate new concepts to replace the less useful ones from the previous iteration. Using the empirical results as feedback, we ensure that our CM remains adaptable and up-to-date with the most relevant molecular concepts. Through this iterative refinement process, we guarantee that the model performance improves over iterations and prune the irrelevant concepts generated in previous iterations. The prompt for this step is shown in Appendix A: Figure~\ref{figure:llm_prompt}.

\section{Experiments}
\label{sec:experiment}

\subsection{Experiment settings} \label{sec:experiment_settings}
\textbf{Datasets } 
We include four datasets from MoleculeNet~\cite{wu2018moleculenet}: two regression datasets (FreeSolv and ESOL) and two classification datasets (BBBP and BACE). FreeSolv provides hydration-free energy for 642 molecules, while ESOL contains water solubility data for 1128 small organic molecules. BBBP (2,039 molecules) assesses blood-brain barrier penetration, and BACE (1,513 molecules) predicts $\beta$-secretase 1 inhibitors for Alzheimer's research. All datasets use the scaffold splits from the Open Graph Benchmark (OGB)~\cite{hu2020open}. Additionally, we include two HTE datasets, Buchwald-Hartwig (BH)~\cite{bhpaper} and Suzuki-Miyaura (SM)~\cite{suzuki}, for reaction yield prediction, with BH covering 3,957 molecules and SM covering 5,650. These datasets use the same splits as~\cite{guo2023can}.


\textbf{Metrics } We follow the standard evaluation metrics for these datasets. For FreeSolv and ESOL, results are measured with Root Mean Square Error (RMSE). For BBBP and BACE, we mainly evaluate these datasets using AUC-ROC and report results in our main Table~\ref{table:mol_main}. Since~\cite{guo2023can} evaluates them with accuracy, we also report accuracy comparison in Table~\ref{table:bace_bbbp_acc} in Appendix~\ref{app:bace_bbbp_acc}. For BH and SM, we evaluate with accuracy.

\textbf{Baselines } Our baselines include GNNs, LLM ICL, and GNN + CBM. Specifically, we use the GIN and GCN. For LLM ICL, we refer to the findings in~\cite{guo2023can} and use their prompts. For GNN + CBM, we use GIN and use GPT-3.5 Turbo for generating concept labels for CBM. For HTE datasets, we only consider LLM ICL baselines as graphs are not provided for the test set.

\textbf{Models } We employ GPT-3.5 Turbo as our primary LLMs for generating concepts and direct labeling. Additionally, we utilize GPT-4 for labeling strategy 2: function code generation, and strategy 3: external tool calling. For strategy 3, the LLM will create code snippets for invoking RDKit~\cite{rdkit}. We don't use GPT-4 for direct labeling due to the high cost of per-instance labeling. After collecting the concept labels, we explore four types of prediction models to cover a broad spectrum of tasks and performance levels. As a basic setting, we use linear models like linear regression and logistic regression, we also consider more advanced models including decision trees and 2-layer MLPs. We use off-the-shelf prediction models from sklearn~\cite{scikit-learn}. We do ablation on LLMs with Claude-2 in Appendix~\ref{app:ablation}. Since we call LLMs through their APIs and the prediction models are light and off-the-shelf, there is no specially requirements, like GPUs, for our framework.

\textbf{Concept Selection } 
We employ AIC~\cite{akaike1973information, akaike1974new} for regression and RFE~\cite{Guyon2002GeneSelection} for classification. These selection methods are specifically applied to linear models, and we use the selection results for multi-iteration performance with decision trees and MLPs. 

\subsection{\method-induced CM performance}
\renewcommand{\thefootnote}{\arabic{footnote}}
\begin{table*}
  \centering
  \scalebox{0.9}{
  \begin{tabular}{l|cccccc}
    \toprule
    & FreeSolv ($\downarrow$) & ESOL ($\downarrow$) & BBBP ($\uparrow$) & BACE ($\uparrow$) & BH ($\uparrow$) & SM ($\uparrow$) \\
    \midrule
    GIN & 2.151 &0.998 & \textbf{69.710} &  \textbf{73.460} & - & - \\
    GCN & 2.186 & 1.015 & 67.800 & 68.930 & - & - \\
    \midrule
    GIN + CBM & 2.412 & 1.373 & 54.500 & 68.457 & - & - \\
    \midrule
    GPT-3.5 Turbo (zero-shot) & 5.450 & 2.039 & 49.256 & 48.765 & 0.320 & 0.473 \\
    GPT-3.5 Turbo (4-shot) & 4.852 & 1.161 & 51.580 & 41.871 & 0.640 & 0.630 \\
    GPT-3.5 Turbo (8-shot) & 4.491 & 1.128 & 56.632 & 47.757 &  0.706 & 0.693 \\
    \midrule
    \method-CM (ours) &  \textbf{2.065} & \textbf{0.843} &  65.278 & 70.744 &  \textbf{0.810} & \textbf{0.800} \\
    \bottomrule
  \end{tabular}
  }
  \caption{Performance comparison of the \method-induced CM with baselines. MoleculeNet regression tasks (FreeSolv and ESOL) are measured in RMSE ($\downarrow$). MoleculeNet classification tasks (BBBP and BACE) are measured in AUC-ROC ($\uparrow$). HTE datasets (BH and SM)\protect\footnotemark are measured in accuracy ($\uparrow$). Ours achieve better results on MoleculeNet regression and HTE tasks and competitive results on MoleculeNet classification tasks.}
  \label{table:mol_main}
\end{table*}


In Table~\ref{table:mol_main}, we compare the performance of the \method-induced CM to baselines. Compared to GNNs, our CM achieves better results on MoleculeNet regression tasks and HTE tasks and competitive results on MoleculeNet classification tasks. In comparison to the results presented by ICL, our models have demonstrated a substantial performance advantage on all tasks. Our best-performing model is the culmination of multiple iterations of refinement and a combination of labeling strategies. Specifically, the results presented in Table~\ref{table:mol_main} are achieved using the following approaches: \textbf{1.} A combination of all three labeling strategies for concept labeling, with further details provided in Appendix \ref{app:ablation_combine}; \textbf{2.} The optimal CMs from linear models, decision trees, and MLPs; \textbf{3.} Concepts refinement over three iterations, as discussed in Section \ref{sec:RQ4}. An in-depth exposition of these techniques is discussed in detail in the RQs below through experiments on MoleculeNet datasets. More details on experiment results for HTE datasets are shown in Appendix~\ref{app:BH_SM}.

\subsection{RQ1: Can \method\ generate meaningful molecular concepts?}

\begin{figure*}[t]
\centering
\includegraphics[width=\textwidth]{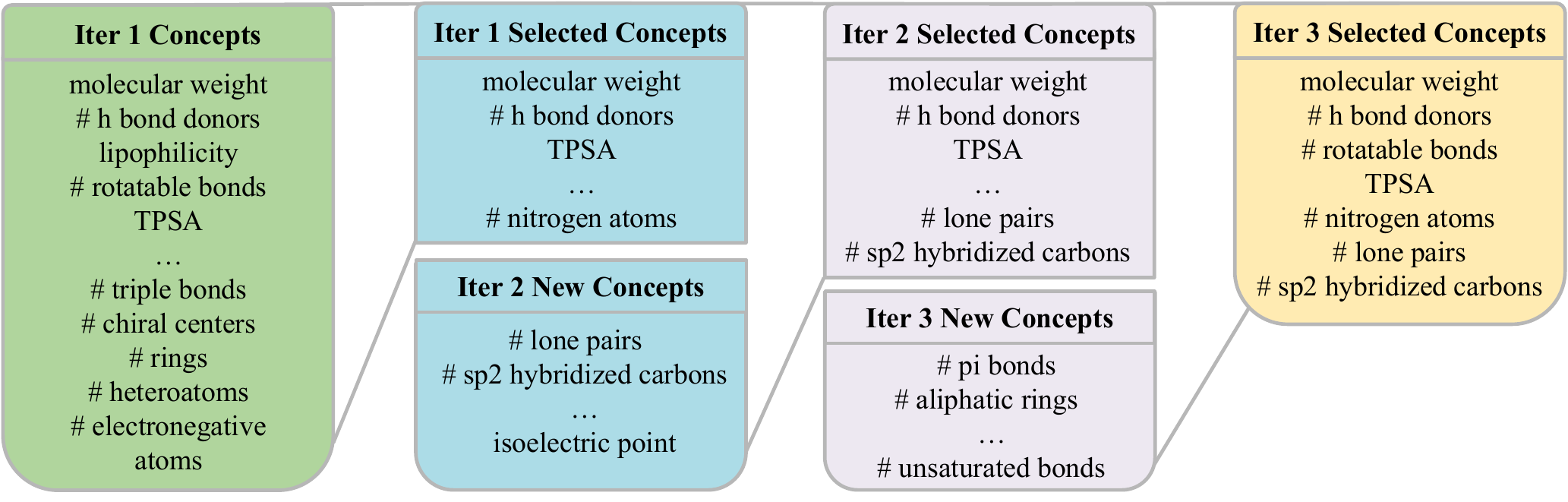}
\caption{RQ1: Concepts selected by \method\ in three refinement iterations on FreeSolv. A detailed version in Appendix~\ref{app:concept_refine_full} Figure~\ref{figure:concept_refine_full}.
}
\vskip -0.2in
\label{figure:feat-iter}
\end{figure*}

The effectiveness of the CM relies on meaningful concepts, traditionally provided by domain experts~\cite{koh2020concept}. In RQ1, we evaluate \method's concept generation through iterative refinement and expert consultation. Figure~\ref{figure:feat-iter} illustrates how concepts selected by a linear regression model for predicting solubility (FreeSolv) evolve from a broad initial set to a more focused, chemically relevant set. Experts noted that eliminated concepts, such as counts of specific atoms and rotatable bonds, while informative, contributed less to predictive power or were correlated with retained concepts. The selected concepts, including molecular weight, \# hydrogen bond donors, and TPSA, are fundamental properties influencing molecular interactions and solubility. For example, hydrogen bond donors relate directly to hydrogen bonding, and TPSA quantifies polar interaction surfaces crucial for solubility in polar solvents~\cite{pajouhesh2005}. These final concepts align well with domain knowledge.

\footnotetext{The performance of GNN-based methods on BH and SM is not reported because we could not obtain the correct train and test splits for their graph datasets from either the original paper or subsequent papers that utilized BH and SM.}

\method\ also mitigate the potential LLM hallucination issue when generating concepts. Our experiments show LLMs perform better on straightforward, well-studied tasks but struggle with complex, experimental, or sparsely documented topics. However, in both cases, LLM hallucination and framework performance can be mitigated by employing combined labeling strategies and iterative refinement, resulting in high quality molecular concepts.

\subsection{RQ2: Can \method\ assign molecules reasonable concept labels using each strategy?}

Accurate concept labels are another critical component of CM performance. In this RQ2, we evaluate 
\method\ labeling results. We collect ground truth labels for concepts where labels are available, either through calculation (e.g., for molecular weights), or manual lookup (e.g., for melting points). We evaluate labels produced by our direct prompting strategy and function generation labeling strategy using the Pearson correlation coefficient ($r$) with the ground truth, due to the scale-invariant nature of the metric. The external tool calling strategy is excluded from this evaluation as tools will always provide correct labels, and the downside of this strategy is that not all the concepts are tool-calculable (e.g., melting points). Results in Table~\ref{table:rq2} show strong correlations can be achieved on most datasets with \method\ labeling. Nonetheless, variations in correlation underscore the potential for method improvement. 

It's worth notice that correlations in Table~\ref{table:rq2} do not have a direct connection to the performance of each labeling strategy. It serves as a sanity check for labeling strategies but an incomplete picture of prediction performance. This is because some concept correlations are non-computable due to missing ground-truth labels, which also contribute to the prediction performance results in Table~\ref{table:RQ3}.

In addition to this benchmarking effort, we also discuss several challenges we encountered and overcame in each labeling strategy, including imputation for missing values, dictionary for unit inconsistency, and Chain-of-Thoughts (CoT) prompts for syntax errors in function code: 

\paragraph{Direct LLM prompting}
 For labeling with direct LLM prompting, we encountered two key issues: missing labels and unit inconsistency. 

\textit{Missing labels} 
    One issue we found for concept labeling with direct LLM prompting is that it is challenging to have LLMs generate some concept labels for certain molecules. For instance, LLMs identified \textit{acid dissociation constant (pKa)} as a crucial concept for predicting water solubility. However, pKa is a quantity only apply to acids, and thus the model will output ``Unknown'' for the label. For the ESOL dataset, this results in a $13.03\%$ missing rate for this concept. The missing-label issue underscores \method's limitation in recognizing concept applicability across molecules. To mitigate this, we apply various imputation methods, including mean value imputation and domain-knowledge-driven imputation. For the latter, we set missing \textit{pKa} labels to $100$—significantly above water's \textit{pKa} of $14$—to denote weak or non-acidity, which enhances the CM's performance.

 
\textit{Unit inconsistency}
For concepts with multiple possible units, labels generated by LLMs can exhibit inconsistent units across molecules. For example, in our experiments on the ESOL dataset, the LLM suggested "melting point" as relevant for predicting water solubility but used inconsistent units—Celsius ($^\circ$C), Fahrenheit (F), or Kelvin (K)—for the melting point values. 

Similar issues arose with other concepts like \textit{molecular volume} and \textit{molecular surface area} due to randomness in LLM context generation when processing different data instances. Our initial attempts to fix this by specifying units in the prompt were ineffective. To address this, we introduced an intermediate step: the LLM generates a concept-to-unit dictionary for proposed concepts, which is then integrated into Step 2's prompt to ensure consistent units in the generated labels.


\paragraph{Function code generation}
When generating labeling functions in Python code, we find it is non-trivial to prompt LLMs for executable functions with no errors. We made two efforts to increase the likelihood of producing executable functions with LLMs. We first perform prompt engineering to clearly specify atom types, adjacency matrices, and node and edge features, which enhances the function quality. Through careful prompt engineering, most generated functions for simpler concepts become executable. However, functions for labeling complex concepts like ``number of rings'' are still unlikely to be error-free due to their intricate nature. We thus adopt a chain of thought (CoT) approach to generate functions. For the CoT prompt, we first ask the LLM to describe the function in natural language, which can best leverage the LLM's strength in generating natural language. Then, the CoT prompt asks the LLM to turn the natural language description of the function into Python code, which we found increases the likelihood of generating accurate and executable functions. An example of the CoT function-generation prompt is shown in Figure~\ref{figure:function_generation}.

\paragraph{External tool calling}
Given there are external tools for molecular science with API access, we prompt LLMs to generate code snippets for calling the tool API. We observe that LLMs are adept at obtaining callable APIs for a majority of our generated concepts from step 1, which we successfully employed to calculate the concept labels for each molecule in our dataset. The example prompts and generated API calls can be found in Figure~\ref{figure:external_tool}.
The drawback of this strategy is that the external tool cannot cover all the concepts generated by the LLM, especially for those measured concepts like melting point. For these cases, we turn to the first two strategies for labeling. 


\begin{table*}
  \centering
  \scalebox{0.8}{
  \begin{tabular}{l|l|l|ccccc}
    \toprule
    Labeling Strategy & LLM & Molecule Format & FreeSolv & ESOL & BBBP & BACE \\
    \midrule
    Str-1 Direct Prompt & GPT-3.5 & Name & 0.82 & 0.63 & 0.06 & - \\
    Str-1 Direct Prompt & GPT-3.5 & SMILES & 0.82 & 0.75 & 0.69 & 0.22 \\
    Str-2 Function & GPT-4 &  -  & 1.00 & 0.79 & 0.69 & 0.67\\
    \bottomrule
  \end{tabular}
  }
  \vskip -0.1in
  \caption{RQ2: Percentage of concepts with a high correlation ($r$ score $\geq$ 0.7) with the ground-truth.}
  \label{table:rq2}
  \vskip -0.07in
\end{table*}

\subsection{RQ3: Can \method -generated concepts and labels be utilized to build an effective CM?}

In RQ1, we have verified that the generated concepts are meaningful according to domain experts. In RQ2, we have shown that concept labels are relatively accurately assigned after properly handling potential issues like missing labels and unit inconsistency. In this RQ3, we compare the performance of the \method-induced CMs when different predictions models and labeling strategies are adopted. Results in Table~\ref{table:RQ3} show that \method\ can give reasonable performance even with the most basic direct prompting labeling strategy and the simplest linear model. The good performance of different prediction models demonstrates the quality of the concepts and the effectiveness of \method.

\begin{table*}
  \centering
  \begin{tabular}{l|l|ccccc}
    \toprule
   Labeling Strategy & Prediction Model & FreeSolv($\downarrow$) & ESOL ($\downarrow$) & BBBP ($\uparrow$) & BACE ($\uparrow$) \\
    \midrule
    Str-1 Direct Prompt & Linear/Logistic & 2.685 & 1.250 & 52.836 & 56.894 \\
    Str-1 Direct Prompt & Decision Tree & 2.791 & 1.272 & 56.887 & \textbf{68.632} \\
    Str-1 Direct Prompt & MLP & 2.338 & 1.194 & 51.794 & 60.059 \\
    Str-2 Function  & Linear/Logistic & 3.284 & 1.254 & 55.671 & 56.624\\
    Str-2 Function & Decision Tree & 2.569 & 1.238 & 54.167 & 55.573 \\
    Str-2 Function & MLP & 2.805 & 1.034 & \textbf{58.738} & 56.894\\
    Str-3 Tool  & Linear/Logistic & 3.142 & 1.011 & 57.350 & 63.154 \\
    Str-3 Tool & Decision Tree & 3.750 & 1.027 & 55.903 & 65.658 \\
    Str-3 Tool & MLP & \textbf{1.981} & \textbf{0.911} & 58.449 & 60.772\\
    \bottomrule
  \end{tabular}
  \vskip -0.1in
  \caption{RQ3: Model performance with different labeling strategies and prediction models.}
  \label{table:RQ3}
\end{table*}

\subsection{RQ4: Does iterative refinement boost the performance of \method-induced CM?} \label{sec:RQ4}
\begin{figure*}[t]
  \includegraphics[width=\textwidth]{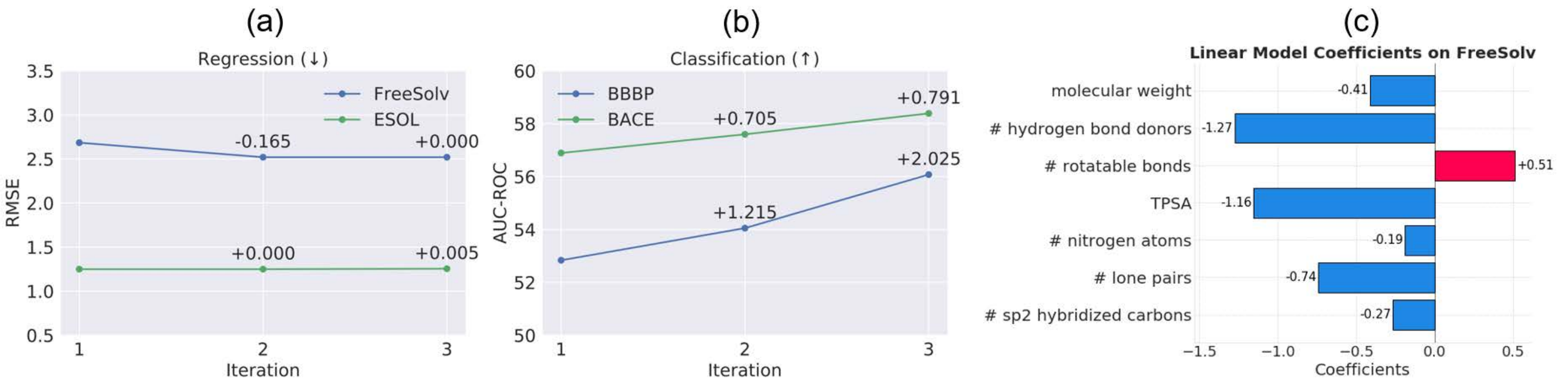}
  \vskip -0.1in
  \caption {RQ4: Iterative refinement improves CM performance for (a) regression and (b) classification tasks; (c) RQ5: Coefficients of the linear regression model from \method\ after 3 iterations on FreeSolv.}
  \label{figure:rq_4_5}
  \vskip -0.1in
\end{figure*}

As one of the most important designs of our \method\ framework, concept refinement helps to identify meaningful important concepts through iterative interactions with LLMs. The concept relevance has been shown to improve in RQ1, but that does not necessarily mean CM performance will also improve. In this RQ4, we run \method\ with three iterative concept refinements on the MoleculeNet datasets with linear prediction models. We show the results in Figure~\ref{figure:rq_4_5} (a) and (b), and we observe that the CM prediction performance indeed improved through concept refinement, especially for classification tasks. The improvement for regression tasks is marginal, partially because the performance is already good for regression.

\subsection{RQ5: Does the \method-induced CM facilitates explainable molecular science?}\label{subsec:rq5}

One of the key advantages of CMs over black-box models is their explainability. In this section, we evaluate this aspect of \method-induced CMs through three experiments on all three types of the prediction models: coefficient interpretation of linear models, split interpretation of decision trees, and concept label intervention of MLPs.

\textbf{Coefficient Interpretation of Linear Models}
Using linears model in the \method-induced CM offers excellent explainability through direct interpretation of the model coefficients. We plot the coefficients of the linear model on FreeSolv for predicting hydration free energy in Figure~\ref{figure:rq_4_5} (c), highlighting three significant concepts: \textit{\# hydrogen bond donors}, \textit{TPSA}, and \textit{\# rotatable bonds}. According to domain experts, the \# hydrogen bond donors relates to a molecule's ability in hydrogen bonding, reflecting its potential to interact with solvents and other molecules. Therefore, an its increment typically leads to a more favorable (more negative) hydration free energy~\cite{h_donor_solubility}. TPSA quantifies the surface area of a molecule that can engage in polar interactions, providing insights into a molecule's permeability characteristics. Thus, higher TPSA also leads to more favorable (more negative) hydration free energy~\cite{pajouhesh2005}. Conversely, the \# rotatable bonds positively correlated with hydration free energy. More rotatable bonds increase molecular flexibility, allowing the molecule to adopt conformations that enhance interactions with water molecules. This increased flexibility can lead to less favorable hydration free energy (less negative), as it reduces the stability of the solvation shell around the molecule~\cite{rotatable_bond_solubility}. Our linear model interpretation aligns with domain knowledge without requiring any human knowledge input into the model. We show results on BBBP in Appendix~\ref{app:rq5}.

\textbf{Splits Interpretation of Decision Trees}
Complementing to the coefficients of the linear model, decision tree enhances the understanding of model's decision process. In Figure~\ref{figure:explain_tree}, we show the 3-layer decision tree for BBBP dataset. In the first two layers, the model uses TPSA to categorize the molecules into four categories, where molecules with TPSA less than 26.99 are likely to penetrate the BBB while molecules with TPSA greater than 176.22 are rarely penetrative. The decision tree further differentiates molecules with TPSA between 26.99 and 107.1 by whether or not it contains a hydrogen bond in its ring structure, where molecules without this property are more likely to penetrate the BBB. On the other hand, the model splits molecules with TPSA between 107.1 and 176.22 using the number of carbon atoms, illustrating that molecules containing more than 23 carbon atoms are very likely to be penetrative. Figure \ref{figure:tree_bbbp} shows the details of the decision tree.

\textbf{Concept Label Intervention}
Besides analyzing interpretable prediction models like linear models and decision trees, we also conduct a case study of concept label interventions with MLPs. Our goal is to identify molecules with similar concept labels except for the one we intend to intervene on (e.g., similar molecular weights, \# aromatic rings, etc., except logP) but different task labels (e.g., soluble vs. insoluble). Two examples we identify from the ESOL dataset are: \textit{Diphenylamine (N(c1ccccc1)c2ccccc2)} and \textit{RTI 17 (CCN2c1ccccc1N(C)C(=S)c3cccnc23)}. After three iterations of refinement these two molecules have the same labels for three out of the four remaining concepts, except for their logP labels, which differ by 0.275 (standardized to have mean 0 and standard deviation 1). Diphenylamine is predicted to be insoluble (-3.648), whereas RTI 17 is predicted to be soluble (-4.079), based on a conventional solubility threshold of -4 \cite{sorkun2019aqsoldb}. These predictions proved to be quite accurate, with the ground truth solubility of these two molecules being -3.857 and -4.227, respectively. By intervening on diphenylamine’s logP value (0.209) to match RTI 17’s logP value (0.484) through interpolation, we observe a linear change in solubility. This study highlights the significant impact of logP on solubility predictions, which is consistent with expert conclusions~\cite{logp_solubility, logp_solubility2}, providing insights beyond black-box models. We show the intervention plot in Appendix~\ref{app:rq5} Figure~\ref{figure:intervetion}.

\subsection{Ablation studies} \label{subsec:limitations}
We conduct ablation studies of \method.
We found that \method\ can perform consistently with different LLMs and is robust to molecule input formats. Also, properly combining the labeling strategies can enhance model performance. These results are in Appendix~\ref{app:ablation}.

\section{Conclusion} 
\label{sec:conclusion}
We propose the \method\ framework that automates the generation and labeling of molecular concepts, overcoming challenges of existing CMs and enhancing explainability through iterative refinement of useful concepts. We demonstrate that, for molecular property prediction tasks, simple linear prediction model on our generated concepts can perform competitively or even better than GNNs and LLM ICL. Our work paves the way for future research to further exploit the capabilities of LLMs for XAI in molecular science and beyond.


\section{Limitations}
The \method\ framework's performance and explainability rely on the quality of LLM-generated concepts and labels. Although we conducted rigorous experiments and verified the LLM-generated concepts and labels with domain experts to ensure the quality of our experiment results, limitations of LLM, such as potential hallucination and rare occurrence of certain chemical formula representations during pre-training stage, may effect the performance and reliability of our framework in other instances. 

As a general framework, we expect the performance and reliability of \method\ to be further improved with newer and more advanced LLMs. A thorough investigation on mechanistic interpretability of LLM and a more powerful LLM dedicated to molecular science may address these issues and could be considered as future directions. 

Another limitation is that the evaluation of the generated concepts and labels often requires validation by human experts, introducing subjectivity and dependency on domain knowledge. Developing automated evaluation methods is another potential direction for improvement.

\bibliography{custom}

\clearpage
\appendix
\section*{Appendices}
\section{Example prompts}\label{app:prompts} 

\begin{figure*}[hbt!]
\centering
\includegraphics[width=\textwidth]{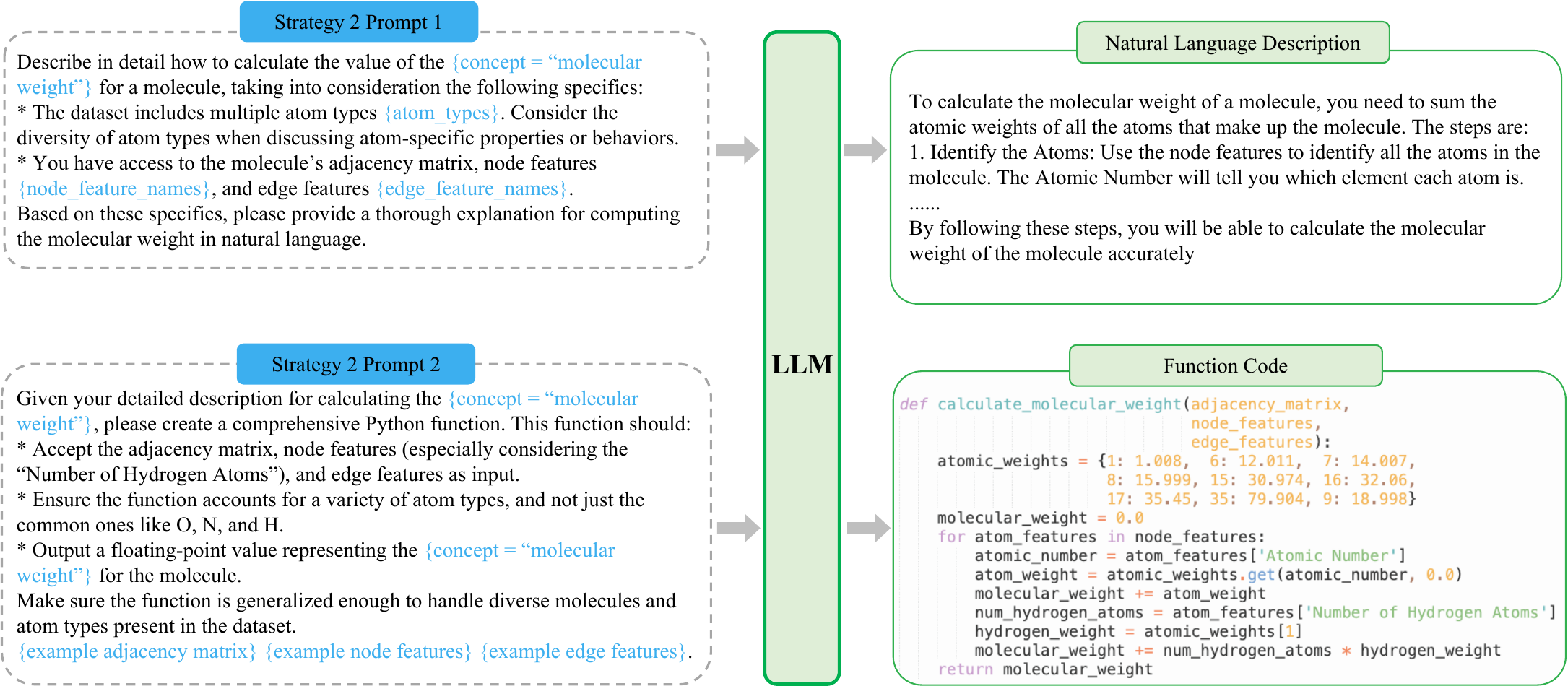} 
\caption{Prompts for generating concept labeling functions in Python code for the FreeSolv dataset.}
\label{figure:function_generation}
\end{figure*}

\begin{figure*}[ht]
\centering
\includegraphics[width=\textwidth]{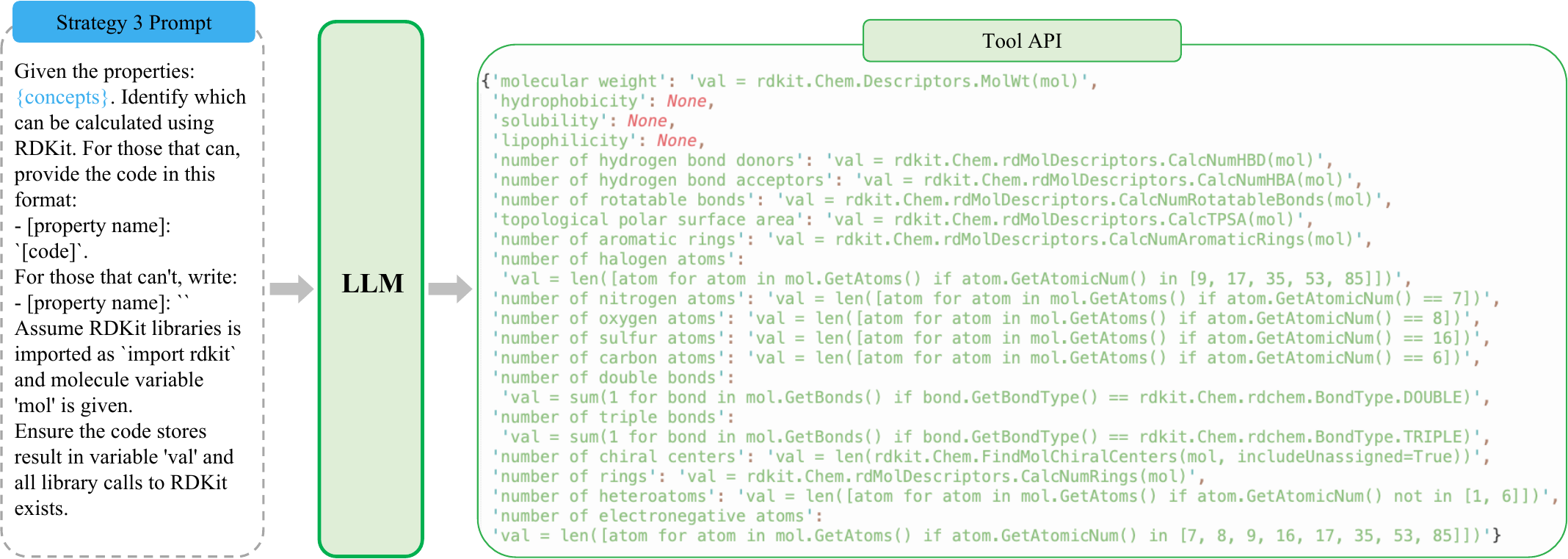} 
\caption{Prompts for calling the external tool RDKit to label concepts on the FreeSolv dataset.}
\label{figure:external_tool}
\end{figure*}

We show an example prompt for generating the labeling functions in Python code in Figure~\ref{figure:function_generation} and an example prompt for generating code snippet to call external tools in Figure~\ref{figure:external_tool}. 

\begin{figure*}[t]
\centering
\includegraphics[width=\textwidth]{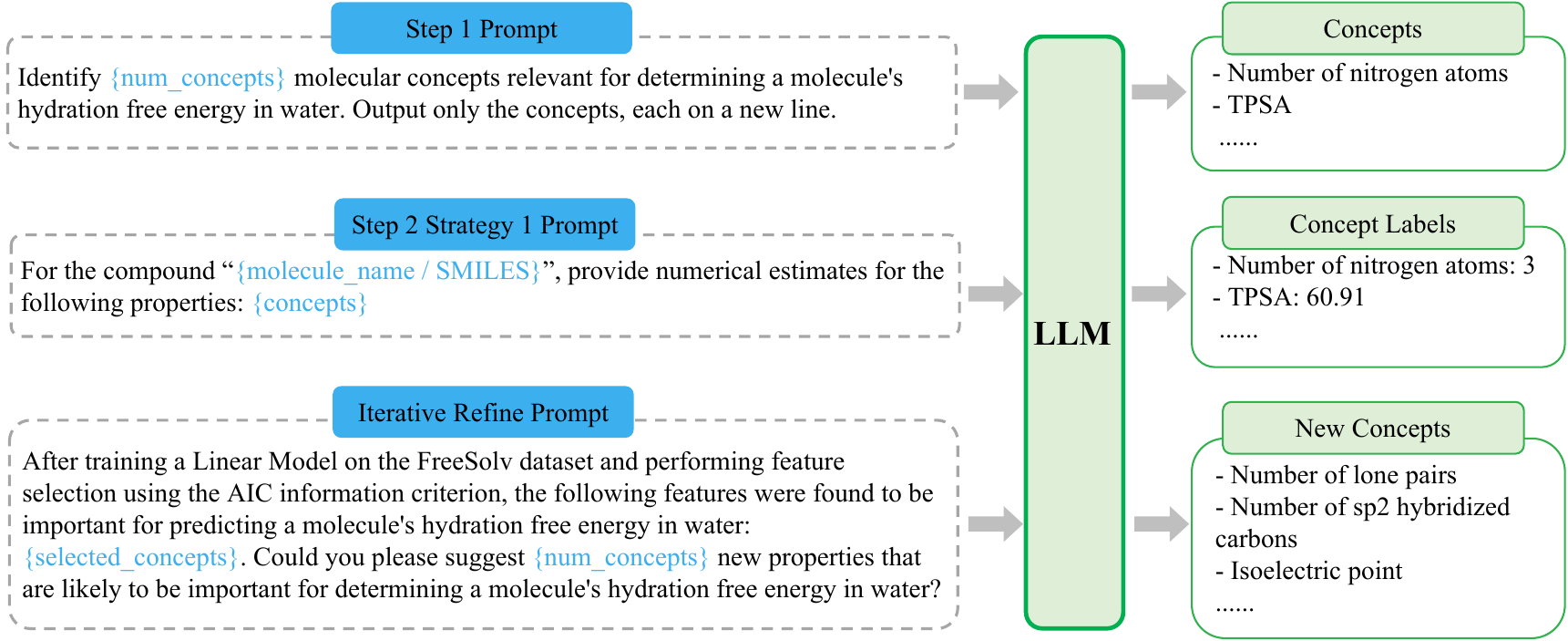} 
\caption[Prompts for concept generation and labeling on the FreeSolv dataset.]{Prompts for concept generation and labeling on the FreeSolv dataset. Hyperparameters, molecule instance information, and re-used LLM responses from a previous step are in blue.}
\label{figure:llm_prompt}
\end{figure*}

We also show Prompts for concept generation and labeling on the FreeSolv dataset for GPT 3.5-Turbo in Figure~\ref{figure:llm_prompt}.

\section{RQ1 supplement: the full version of concept refinement}\label{app:concept_refine_full}

\begin{figure*}[hbt]
\centering
\includegraphics[width=0.9\textwidth]{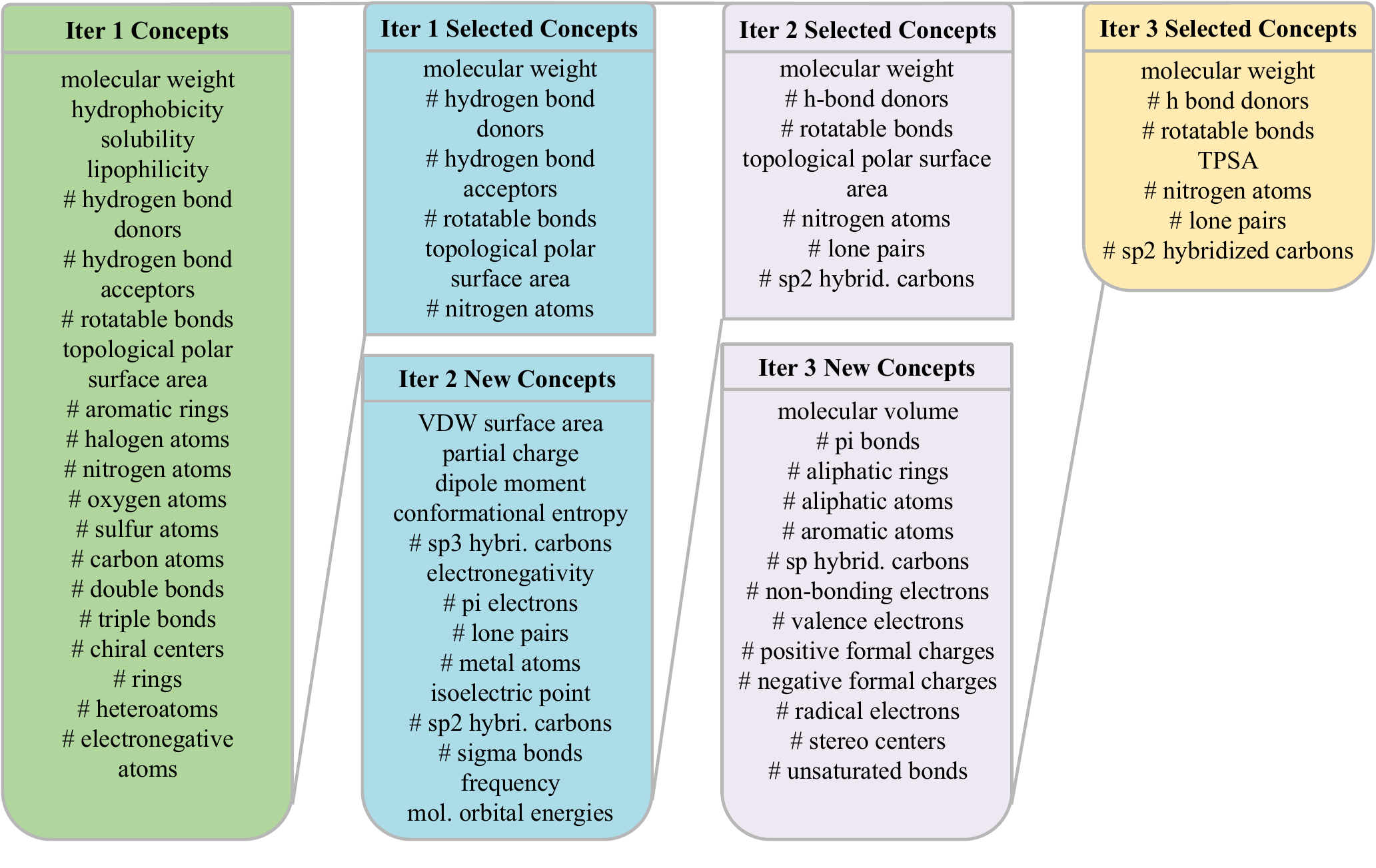} 
\caption{Concepts selected by \method\ in three refinement iterations on FreeSolv. Full version.}
\label{figure:concept_refine_full}
\end{figure*}

In Figure~\ref{figure:concept_refine_full}, we show the full list of concepts selected by \method on FreeSolv in three refinement iterations. The result corresponds to the RQ1.

\section{RQ5 supplement: BBBP linear model coefficients and ESOL intervention visualization} \label{app:rq5}

\begin{figure}[t]
\centering
\includegraphics[width=\columnwidth]{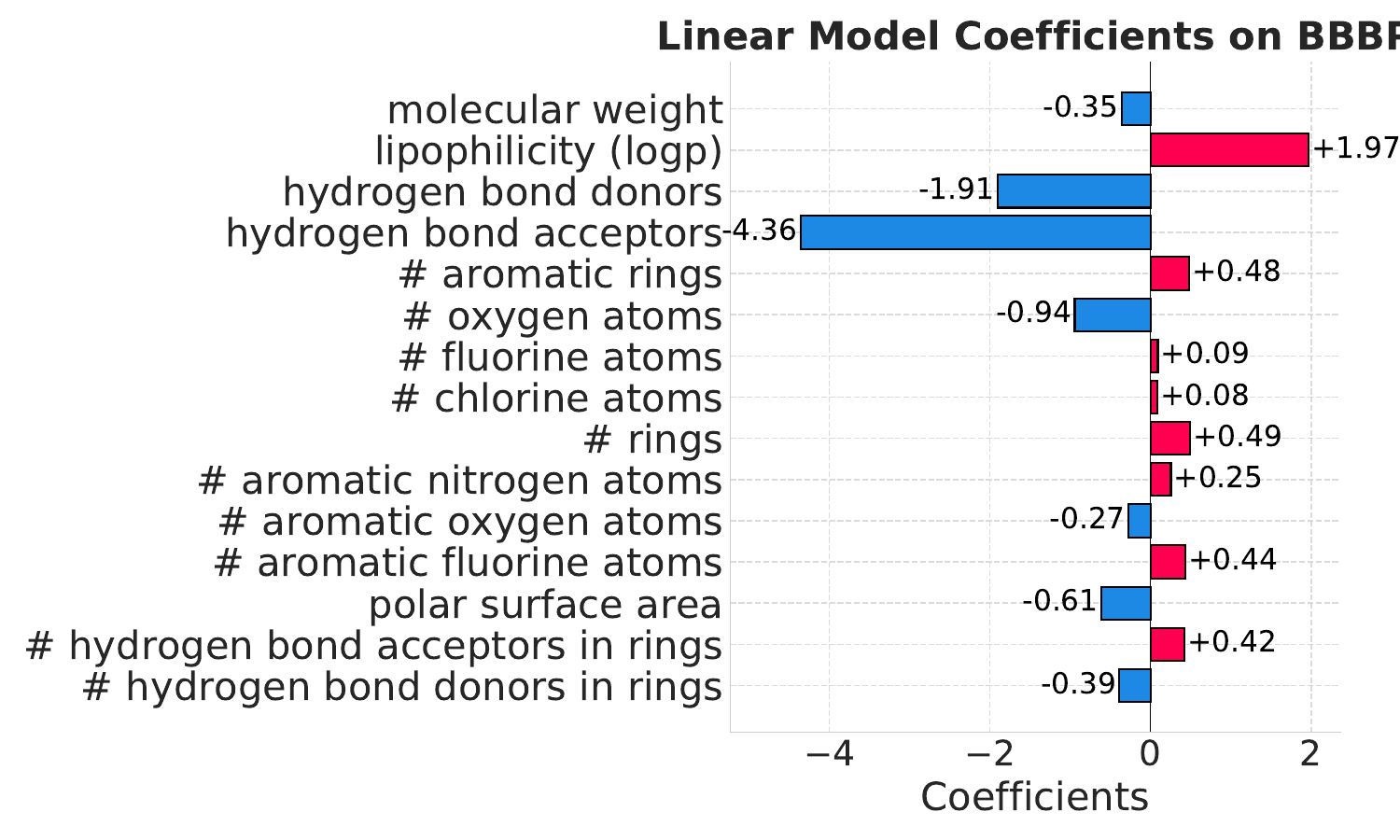} 
\caption{RQ5: Coefficients of the logistic regression model on BBBP with concepts refined by \method\ after three iterations}
\label{figure:explain_linear_bbbp}
\end{figure}

\textbf{Coefficient Interpretation of Linear Models}
Additional to the results in Section~\ref{subsec:rq5}, we evaluating the linear model on the BBBP dataset. We focused on the top positive coefficient lipophilicity (logP) and the top negative coefficient hydrogen bond acceptors shown in Figure~\ref{figure:explain_linear_bbbp}. Notably, logP has a coefficient of 1.97 and  
hydrogen bond acceptors has a coefficient -4.36. These finding aligns with domain knowledge, as higher logP enhances a molecule's ability to cross lipid-rich biological membranes. Conversely, a lower number of hydrogen bond acceptors generally enhances a molecule's permeability through the BBB. These findings validate the CM’s alignment with established biochemical principles, demonstrating its potential utility in predictive modeling for molecular properties.

\begin{figure}[t]
  \includegraphics[width=\columnwidth]{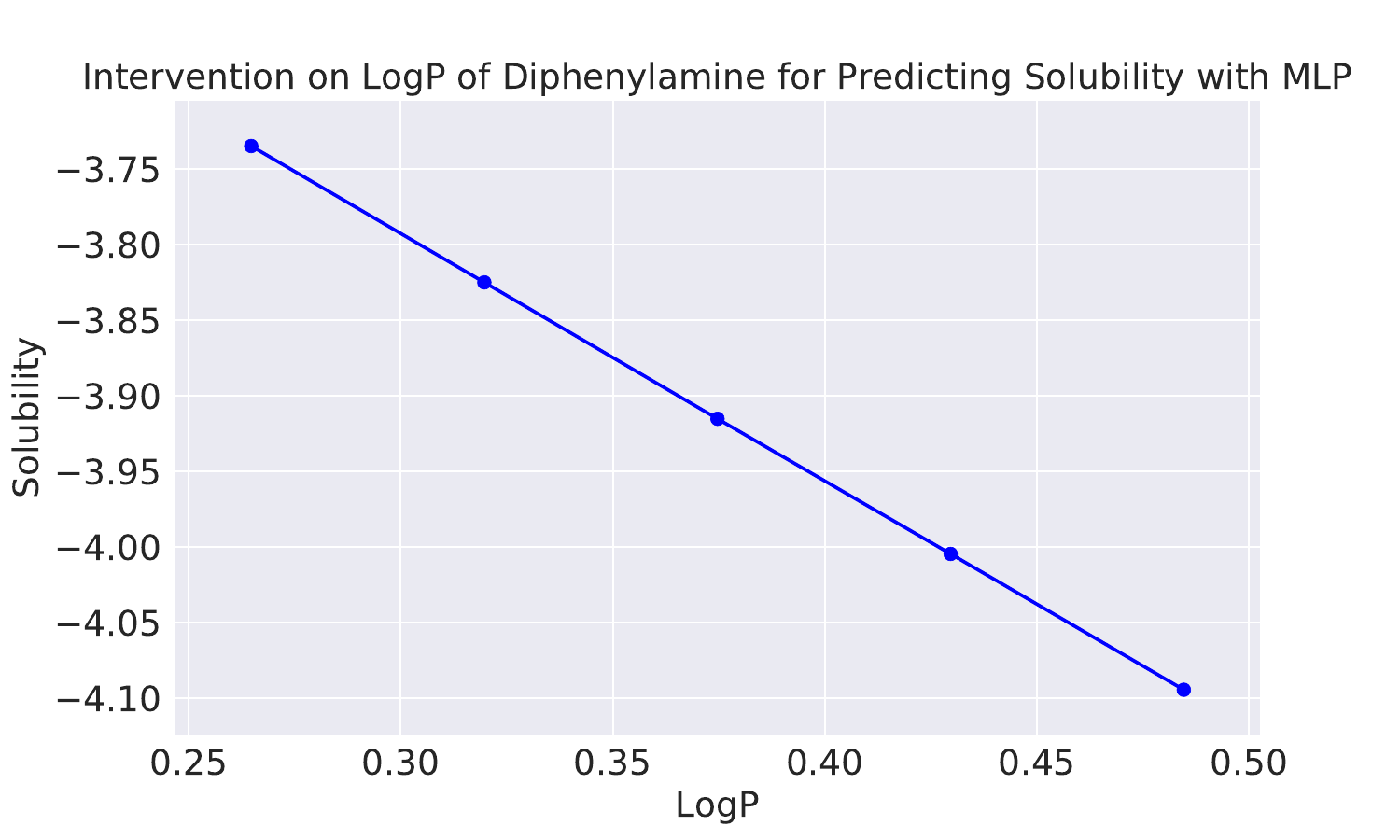}
  \caption{Intervention on logP of diphenylamine for predicting solubility with MLP}
  \label{figure:intervetion}
\end{figure}

\textbf{Concept Label Intervention} Complementing to the intervention study in Section~\ref{subsec:rq5}, we plot the results in Figure~\ref{figure:intervetion}.

\begin{figure*}[t]
  \centering
  \includegraphics[width=0.8\textwidth]{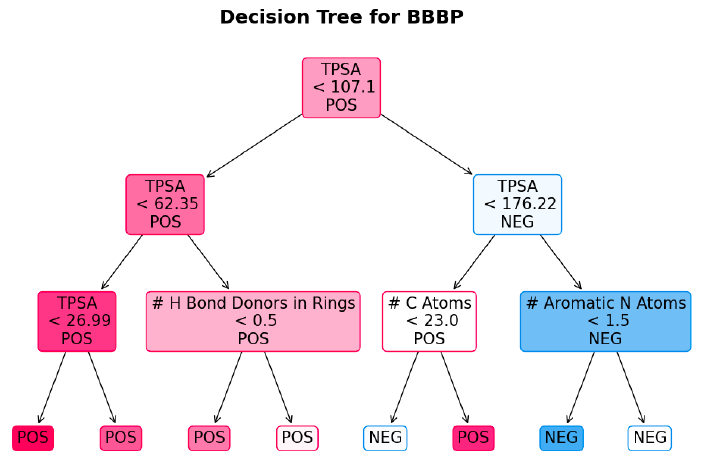}
  \caption{RQ5: coefficients of decision tree on BBBP from \method\ after three iterations.}
  \label{figure:explain_tree}
\end{figure*}

\textbf{Coefficients of decision tree} Complementing to the intervention study in Section~\ref{subsec:rq5}, we plot the coefficients of decision tree in Figure~\ref{figure:explain_tree}.

\section{Ablation Studies} \label{app:ablation}

\subsection{Different LLMs}
We study the performance of \method\ with different LLMs. Table~\ref{table:ablation_llm} compares the performance of GPT-3.5 Turbo and Claude-2 using the direct LLM prompting labeling strategy with linear prediction models. While both GPT-3.5 Turbo and Claude-2 exhibit slightly inferior performance compared to GNNs across four datasets, they maintain competitive results, emphasizing simplicity and interpretability. Specifically, Claude-2 underperforms GPT-3.5 Turbo after first iteration, potentially due to its less consistent and accurate response. This inconsistency, partly attributed to more frequent issues with missing values and unit inconsistencies observed in Claude-2, suggests GPT-3.5 Turbo's superior ability to generate reliable ground truth knowledge. Additionally, GPT-3.5 Turbo's better prompt comprehension and domain knowledge in chemistry might contribute to its enhanced performance in predicting target concepts.

\begin{table}
  \centering
  \resizebox{\columnwidth}{!}{%
  \begin{tabular}{l|cccc}
    \toprule
    LLM & FreeSolv($\downarrow$) & ESOL ($\downarrow$) & BBBP ($\uparrow$) & BACE ($\uparrow$) \\
    \midrule
    GPT-3.5  & 2.685 & 1.250 & 52.84 & 56.89 \\
    Claude-2 & 2.804 & 1.327 & 52.78 & 56.11 \\
    \bottomrule
  \end{tabular}
  }
  \caption{Ablation on LLMs (GPT vs. Claude-2).}
  \label{table:ablation_llm}
\end{table}

\subsection{Direct LLM prompting with molecule names vs. with SMILES strings}\label{app:SMILES_vs_names}


Building on insights from ~\cite{guo2023can} regarding LLMs' challenges with long molecular representations, we examined LLM's capability in labeling concepts using SMILES strings and molecule names across our datasets. Our findings indicate that LLMs perform reasonably well in identifying basic concepts like molecular weight and atom counts using either molecule names or SMILES strings, with a strong correlation to ground truth labels ($r > 0.9$). However, LLMs struggle with complex concepts requiring detailed structural knowledge, such as the \textit{number of chiral centers}. Moreover, our analysis reveals a notable decline in LLM's performance with molecule names in the larger datasets like BBBP, suggesting LLM's familiarity with common molecular names improves its performance on smaller datasets, but this advantage diminishes with less familiar names in larger datasets. In contrast, the structural specificity of SMILES strings maintains more consistent performance across dataset sizes, highlighting their utility in representing unique molecular concepts. Furthermore, we compared the performance difference between the two representations over 3 iterations. As demonstrated in Table~\ref{table:smiles_vs_names_iter}, the model performance using SMILES strings matched the model performance using molecule names on most datasets, but a notable improvement in performance with SMILES strings is observed on the BBBP dataset.

\begin{table*}
  \centering
  \begin{tabular}{l|cc|cc|cc|c}
    \toprule
    & \multicolumn{2}{c|}{FreeSolv ($\downarrow$)} & \multicolumn{2}{c|}{ESOL ($\downarrow$)} & \multicolumn{2}{c|}{BBBP ($\uparrow$)} & BACE ($\uparrow$) \\
 Input Format & SMILES & Names & SMILES & Names & SMILES & Names & SMILES \\
    \midrule
    GPT-3.5 iter 1 & 2.854 & 2.685 & 1.401 & 1.250 & 53.88 & 52.84 & 56.89 \\
    GPT-3.5 iter 2 & 2.662 & 2.520 & 1.262 & 1.250 & 56.08 & 54.05 & 57.60 \\
    GPT-3.5 iter 3 & 2.763 & 2.520 & 1.262 & 1.255 & 60.41 & 56.08 & 58.38 \\
    \bottomrule
  \end{tabular}
  \caption{Ablation on input formats (SMILES vs. molecule names).}
  \label{table:smiles_vs_names_iter}
\end{table*}

\begin{figure*}[ht]
  \centering
  \includegraphics[width=\textwidth]{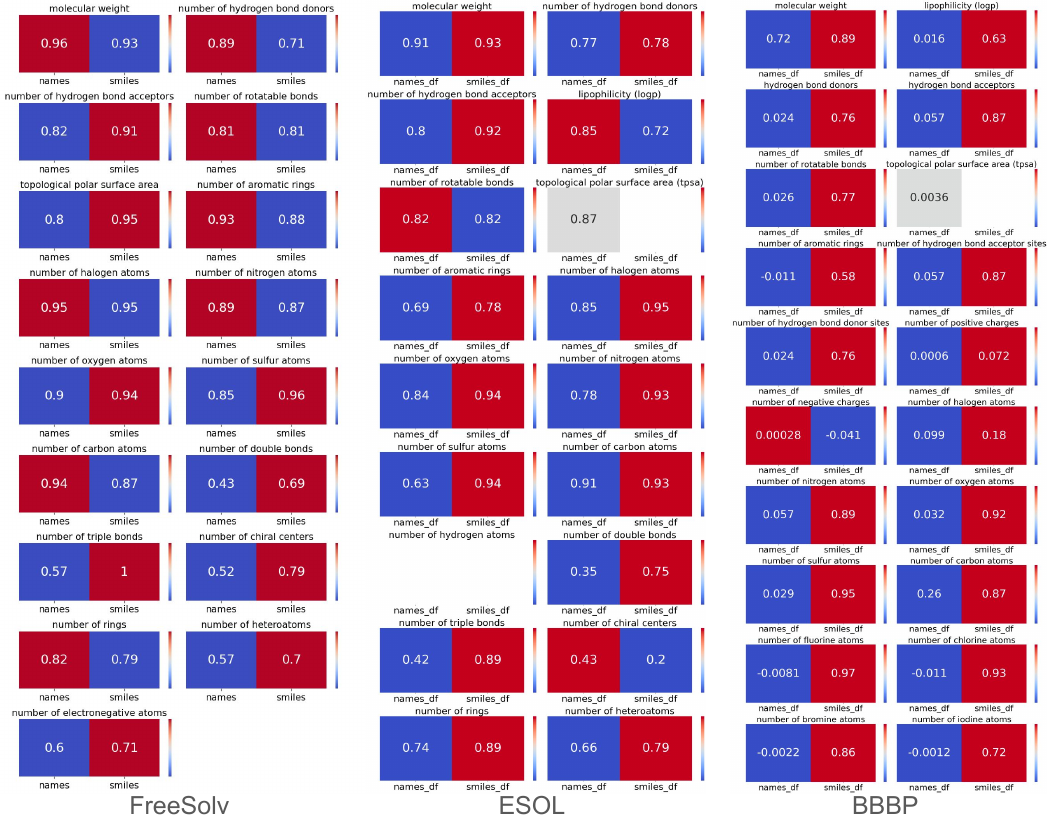} 
  \caption{Correlation ($r$) between the ground truth labels and concept labels generated using molecule names or SMILES strings. Red indicates a higher correlation.}
  \label{figure:SMILESvsnames}
  \end{figure*}

We also present a comparative analysis of the quality of concept labels generated using molecular names vs. SMILES strings. The comparison is visualized through a series of heatmaps, as illustrated in Figure~\ref{figure:SMILESvsnames}.


\subsection{Combine different labeling strategies} \label{app:ablation_combine}

The \method\ framework includes three labeling strategies and allows easy extension to new ones. We consider combinations of the labeling strategies and study their impact on model performance, where we adopt a simple priority heuristic where strategy 3 $>$ strategy 2 = strategy 1. Specifically, whenever the external tool is available for a suggested concept, we get the accurate concept labels from calling it. Otherwise, two concept labels are derived from both direct prompting and function code generation, and they are both considered in step 3 for the selection. This combined-strategy labeling turns out to outperform most of the standalone strategies as shown in Table~\ref{table:ablation_labeling}. These findings demonstrate that the three labeling strategies have their own strengths and weaknesses for different concepts, and they can be complement to each other to maximize the model performance. We leave exploration of new strategies and more sophisticated strategy combinations as future work.


\section{Accuracy comparison with LLM ICL on BBBP and BACE} \label{app:bace_bbbp_acc}

In our experiments, we follow the standard and widely-used evaluation metrics for all datasets. For classification tasks on BBBP and BACE, we mainly evaluate these datasets using the AUC-ROC metric and report results in our main Table~\ref{table:mol_main}. Since~\cite{guo2023can} provides the ICL prompts but evaluates BBBP and BACE with accuracy, we also report a comparison in accuracy in Table~\ref{table:bace_bbbp_acc} for a fair comparison.

\section{Decision trees visualization} \label{app:}
\begin{figure*}[h]
  \centering
  \includegraphics[width=\textwidth]{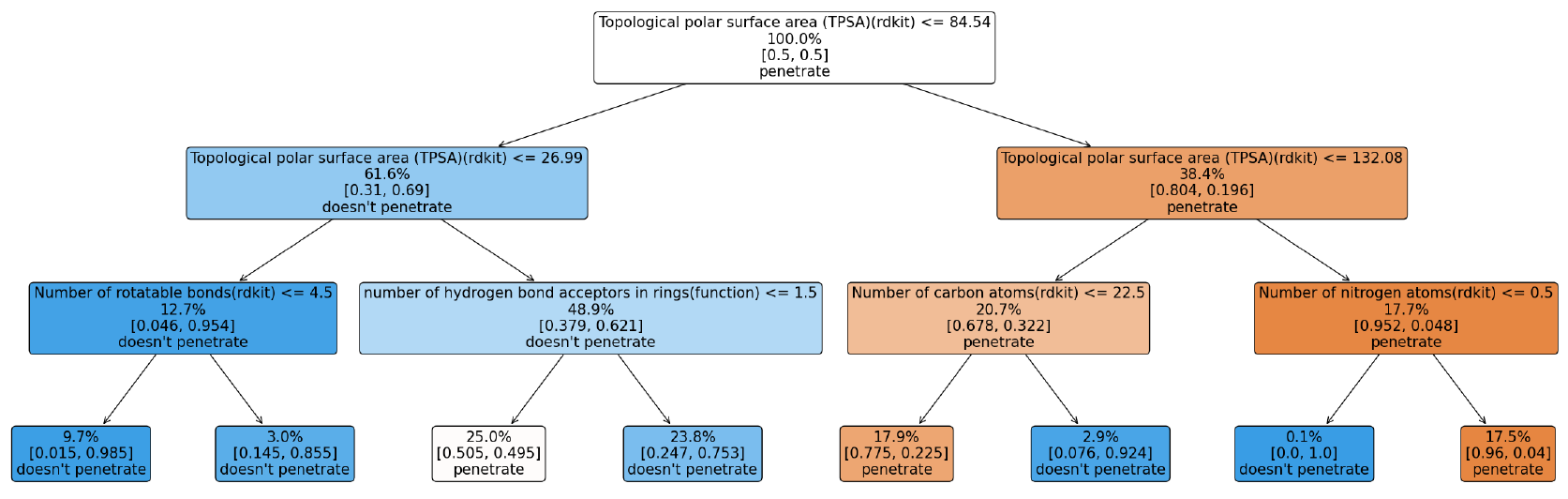} 
  \caption{The decision trees for classification on BBBP.}
  \label{figure:tree_bbbp}
  \end{figure*}

\begin{figure*}[t]
  \centering
  \includegraphics[width=\textwidth]{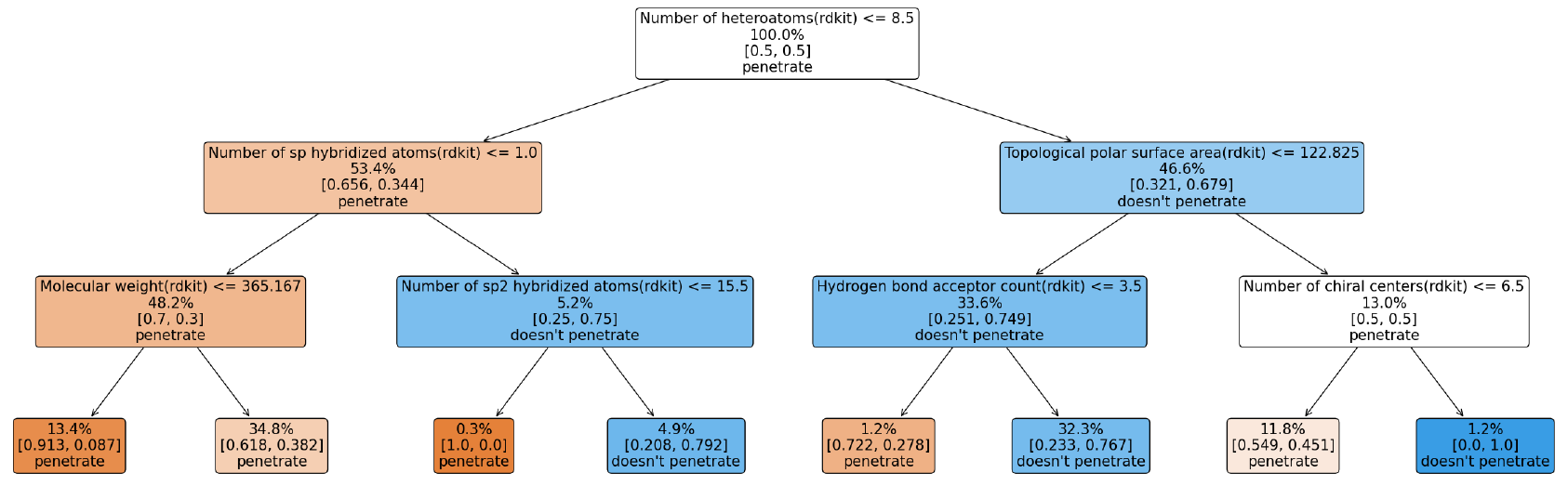} 
  \caption{The decision trees for classification on BACE.}
  \label{figure:tree_bace}
  \end{figure*}

As discussed in \ref{subsec:rq5}, we visualize the decision tree which makes the prediction process explainable. Figure~\ref{figure:tree_bbbp} shows the impurity details of the decision tree shown in Figure~\ref{figure:explain_tree} and Figure~\ref{figure:tree_bace} shows a sample decision tree for the BACE dataset.

\section{More results on Buchwald-Hartwig and Suzuki-Miyaura}\label{app:BH_SM}
The GPT ICL performance from~\cite{guo2023can} are measured on 100 data samples. To compare \method's performance with their numbers, for each dataset we picked the best model performance from the logistic regression models or MLP models trained on either 200 or 500 sampled training data. The performance details are presented in Table~\ref{table:BH_SM_best_acc} with best performance reported in Table~\ref{table:mol_main}.

\begin{table}[th]
  \centering
  \begin{tabular}{l|c|c}
  \toprule
      & BBBP ($\uparrow$) & BACE ($\uparrow$)\\
  \midrule
     GPT-4 (zero-shot)&  0.476 &  0.499 \\
   GPT-4 (Scaffold, k= 8)& 0.614 & 0.679\\
     GPT-3.5 (Scaffold, k= 8) & 0.463 & 0.496 \\ \midrule
     Ours & \textbf{0.657} & \textbf{0.704} \\
  \bottomrule
  \end{tabular}
  \caption{
  Performance comparison of the \method-induced CM vs. LLM ICL (results are taken from~\cite{guo2023can}). Results are evaluated with Accuracy ($\uparrow$).}
  \label{table:bace_bbbp_acc}
  \end{table}

\begin{table}[th]
\centering
\begin{tabular}{l|c|c}
\toprule
    & BH ($\uparrow$) & SM ($\uparrow$) \\
\midrule
   GPT-4 (random, k = 8)&  0.800 &  0.764 \\ \midrule
 GPT-3.5 + logistic (200 samples)& 0.800 & 0.770\\
   GPT-3.5 + MLP (200 samples) & 0.800 & 0.780 \\
   GPT-3.5 + logistic (500 samples) & 0.790 & 0.780 \\
   GPT-3.5 + MLP (500 samples) & \textbf{0.810} & \textbf{0.800} \\
\bottomrule
\end{tabular}
\caption{
Performance comparison of the best \method-induced CM vs. LLM ICL 
 in accuracy ($\uparrow$) (GPT results are taken from~\cite{guo2023can}).}
\label{table:BH_SM_best_acc}
\end{table}

\begin{table*}
  \centering
  \begin{tabular}{l|l|ccccc}
    \toprule
    Labeling Strategy & Model  & FreeSolv($\downarrow$) & ESOL ($\downarrow$) & BBBP ($\uparrow$) & BACE ($\uparrow$) \\
    \midrule
    Direct Prompt & Linear/Logistic & 2.685 & 1.250 & 52.836 & 56.894 \\
    Direct Prompt & Tree Model & 2.791 & 1.272 & 56.887 & \textbf{68.632} \\
    Direct Prompt & MLP &  2.338 & 1.194 & 51.794 & 60.059 \\
    Direct Prompt + Function & Linear/Logistic & 2.697 & 1.254 & 56.134 & 56.712\\
    Direct Prompt + Function & Tree Model & 2.540 & 1.364 & 55.150 & 59.998 \\
    Direct Prompt + Function & MLP & 2.211 & 0.971 & 57.697 & 65.797\\
    Direct Prompt + Function + Tool & Linear/Logistic & 3.002 & 1.136 & 55.845 & 64.032\\
    Direct Prompt + Function + Tool & Tree Model & 3.752 & 1.107 & 56.250 & 65.658 \\
    Direct Prompt + Function + Tool & MLP &\textbf{2.122} & \textbf{0.791} & \textbf{58.391} & 62.624 \\
    \bottomrule
  \end{tabular}
  \caption{Combine labeling strategies.}
  \label{table:ablation_labeling}
\end{table*}

\end{document}